\pdfoutput=1

\documentclass[11pt]{article}

\usepackage[final]{acl}

\usepackage{times}
\usepackage{latexsym}

\usepackage[T1]{fontenc}

\usepackage[utf8]{inputenc}

\usepackage{microtype}

\usepackage{inconsolata}

\usepackage{graphicx}

\usepackage{array}
\usepackage{multirow}
\usepackage{tabularx}
\usepackage{longtable}

\title{To Know or Not To Know? Analyzing Self-Consistency \\ of Large Language Models under Ambiguity}

\author{
    Anastasiia Sedova$^{1,2*}$ \hspace{10pt} 
    Robert Litschko$^{3,4*}$ \hspace{10pt}  
    Diego Frassinelli$^{3}$ \\ 
    {\bf Benjamin Roth$^{1,5}$} \hspace{10pt}  
    {\bf Barbara Plank$^{3,4}$} \\ 
    $^1$Faculty of Computer Science 
    $^2$ UniVie Doctoral School Computer Science \\ $^5$ Faculty of Philological and Cultural Studies, University of Vienna, Austria \\
    $^3$ MaiNLP, Center for Information and Language Processing, LMU Munich, Germany \\ 
    $^4$ Munich Center for Machine Learning (MCML), Germany \\ 
    \texttt{anastasiia.sedova@univie.ac.at} \hspace{20pt} 
    \texttt{rlitschk@cis.lmu.de}
}

\usepackage{makecell}
\usepackage{dblfloatfix}

\newcolumntype{Y}{>{\centering\arraybackslash}X}
\newcommand{\STAB}[1]{\begin{tabular}{@{}c@{}}#1\end{tabular}}

\begin{document}

\maketitle

\def\thefootnote{*}\footnotetext{Equal contribution}\def\thefootnote{\arabic{footnote}}

\begin{abstract}
One of the major aspects contributing to the striking performance of large language models (LLMs) is the vast amount of factual knowledge accumulated during pre-training. 
Yet, many LLMs suffer from self-inconsistency, which raises doubts about their trustworthiness and reliability.
This paper focuses on entity type ambiguity, analyzing the proficiency and consistency of state-of-the-art LLMs in applying factual knowledge when prompted with ambiguous entities.
To do so, we propose an evaluation protocol that disentangles \textit{knowing} from \textit{applying knowledge}, and test state-of-the-art LLMs on 49 ambiguous entities. %
Our experiments reveal that LLMs struggle with choosing the correct entity reading, achieving an average accuracy of only 85\%, and as low as 75\% with underspecified prompts. 
The results also reveal systematic discrepancies in LLM behavior, showing that while the models may possess knowledge, they struggle to apply it consistently, exhibit biases toward preferred readings, and display self-inconsistencies.
This highlights the need to address entity ambiguity in the future for more trustworthy LLMs.%
\footnote{We make the code and the model outputs available: \texttt{https://github.com/anasedova/ToKnow\_or\_NotToKnow}}
\end{abstract}

\section{Introduction}

Large language models (LLMs) have demonstrated remarkable performance in a variety of natural language processing tasks \cite{touvron2023llama,gpt-4o,llama3}, 
also largely due to the extensive factual knowledge they accumulate during pre-training.
A crucial factor in building trust in models is their capacity to generate  consistent and dependable outputs
that align with their internal knowledge \cite{li2024benchmarking,yu2024kola}.
However, LLMs frequently produce unreliable responses: e.g., when externally provided knowledge conflicts with internal parametric knowledge \cite{pan-etal-2023-attacking,xie2024adaptive}, or when exposed to misinformation during pre-training \cite{zhao2024knowing}. %
Such conflicts
often lead to inconsistencies in model responses, 
reducing LLMs' trustworthiness \cite{litschko-etal-2023-establishing,sun2024trustllm}.

\begin{table*}[t]
    \small
    \hyphenpenalty=10000
    \centering
    \begin{tabularx}{\linewidth}{|>{\hsize=0.0005\hsize}l|>{\hsize=1.0\hsize}X|>{\hsize=0.0005\hsize}c|} 
        \hline
        \multicolumn{1}{|c|}{\textbf{\makecell[l]{Entity Type}}} & \multicolumn{1}{l|} {\textbf{List of Entities}} &
        \multicolumn{1}{c|} {\textbf{\makecell[l]{Entity Property}}} \\ \hline
        {animal} &  Jaguar, Puma, Penguin, Greyhound, Dove, Fox, Lynx & {\makecell[l]{speed}} \\ %
        {fruit} &  Apple, Fig, Mango, Kiwi, Papaya, Orange & {\makecell[l]{color}} \\ %
        {myth} & Amazon, Nike, Midas, Mars, Hermes, Hyperion, Vulcan, Pegasus & {\makecell[l]{gender}} \\ %
        person & Ford, Disney, Tesla, Boeing, Dell, Ferrero, Benetton, Levi Strauss, Versace, Philips & {\makecell[l]{date of birth}}\\ %
        {location} & Amazon, Cisco, Montblanc, Patagonia, Hershey, Nokia, Eagle Creek, Prosper & {\makecell[l]{area in m$^2$}}\\ %
        {abstract} & Triumph, Harmony, Genesis, Vision, Pioneer, Vanguard, Zenith, Allure, Tempo, Fidelity & {\makecell[l]{level of abstractness}}\\ \hline
        {\textit{company}} & \textit{all entities listed above} & {\makecell[l]{\textit{founding year}}} \\  \hline
    \end{tabularx}
    \vspace{-5pt}
    \caption{Overview of ambiguous entities. We use a total of 49 entities belonging to 7 entity types. The entities are chosen such that have at least two readings: the listed \textit{entity type} and \textit{company}. Entity properties are chosen such that the entity type can be uniquely inferred from it.}
    \label{tab:categories}
\end{table*}

\begin{figure}
\includegraphics[scale=0.31]{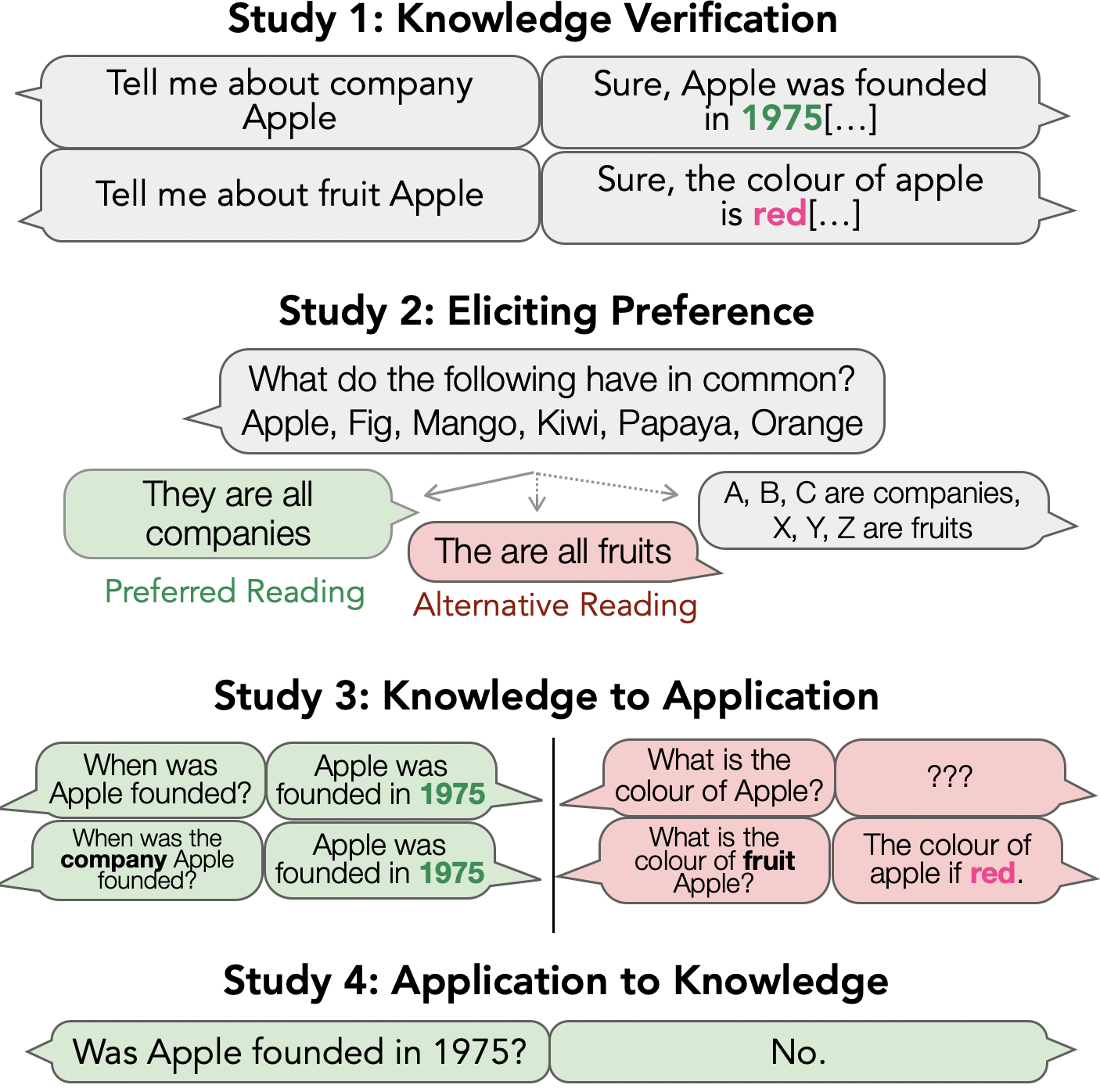}
\vspace{-15pt}
\caption{Overview of our four studies on LLMs' self-consistency using prompts with ambiguous entities. Colors indicate preferred (green) and alternative (red) readings implied in the query or adopted by the model.}
\label{fig:main_figure}
\end{figure}

In this work, we highlight \textit{entity ambiguity}, a frequent challenge for LLMs \cite{liu-etal-2023-afraid,
stengel-eskin2024zero,
kim2024aligning}, as a \emph{source of unreliability} and conduct an in-depth qualitative analysis to better understand how models behave in its presence.
Specifically, we examine the self-consistency\footnote{We analyze consistency on straightforward ``internal knowledge retrieval'' tasks that do not necessitate CoT prompting as in, for example, \citet{wang2023selfconsistency}.}
of state-of-the-art LLMs -- i.e., how well they align with their intrinsic knowledge while avoiding contradictory responses \cite{chen2024two} -- by evaluating their reasoning abilities in contexts involving entity type ambiguity.
Importantly, in our study we provide an operationalization to disentangle LLM's capabilities of \textit{\textbf{K}nowing}\footnote{The term ``knowing'' refers to parametric knowledge as discussed in \citet{mallen2023trust,litschko-etal-2023-establishing}.%
}
(i.e., how aware and sensitive a model is to the possible interpretations, or \textit{readings}, of ambiguous entities), and \textit{ \textbf{A}pplying knowledge} (i.e., how well a model can identify the correct reading when prompted with entity-specific questions and use their parametric knowledge to provide accurate responses about that entity). %
For example, as shown in Figure~\ref{fig:main_figure}, if a model ``knows'' that \textit{apple} can be a fruit and a company, to what extent can we assume that the model infers the company meaning when asked about the founding year of the entity?
Similarly, if a model responds with \textit{``Apple was founded in 1976''} can we assume that it remains self-consistent with its own answer?
Gaining more insights on the capability of an LLM to deal with entity ambiguity has several benefits. It facilitates more natural conversations, as knowing whether an LLM can  
disambiguate an entity allows to minimize the number of clarification questions \citep{
lee-etal-2023-asking}, and enhances trust in LLMs via its ability to acknowledge previously provided knowledge. %
The overarching goal of this work is to provide a behavioral test suite for studying how a model's knowledge of different entity readings interacts with its ability %
to infer and apply the correct reading for a given prompt.

Specifically, we address the following research questions: (\textbf{RQ1}) %
How well do LLMs resolve entity ambiguity within a given prompt context?
(\textbf{RQ2}) How (much) is the ability to infer the correct entity type biased towards “preferred readings”? %
(\textbf{RQ3})
Can LLMs self-verify their answers for entity-related questions, given they have successfully disambiguated them? %

We examine the behavior of six state-of-the-art LLMs, varying in size, type, and whether they are open-source or proprietary: Mistral-7B-Instruct-v0.2 \cite{jiang2023mistral},
Gemma-1.1-7B-IT \cite{gemmateam2024gemma},  Llama-3-70B \cite{llama3}, Mixtral-8x7B-Instruct-v0.1\cite{jiang2024mixtral}, GPT-3.5 \cite{gpt-3.5}, and GPT-4o \cite{gpt-4o}\footnote{Implementation details are provided in Appendix \ref{app:technical_details}.}. Our results show that, despite the seemingly simple task, LLMs fail to  disambiguate and handle entities consistently.

\section{Methodology}
\label{sec:methodology}

To study LLMs' ability to implicitly infer correct entity meanings, we devise a behavioral test suite. We use a set of entities that
can be interpreted as either (1) one of six entity types or (2) company names (Table \ref{tab:categories}), meaning each entity has \textit{at least two} possible interpretations. 
We adopt this approach to distinguish between a preferred and an alternative reading, which allows us to investigate if %
the disambiguation ability of LLMs is consistent or biased across different entity types. 

Our research setup comprises four studies (see Figure~\ref{fig:main_figure}). Study 1 verifies knowledge possession in models; Studies 2 and 3 assess the models' abilities to \textit{apply} this knowledge (K $\rightarrow$ A); and Study 4 evaluates the knowledge possession post-application (A $\rightarrow$ K). Collectively, the results of our four experiments provide us a way to gain knowledge on how LLMs treat entity level ambiguity, i.e., the mutual relationship K~$\leftrightarrow$~A. %

\paragraph{Study 1: Knowledge Verification (K).} First, we analyze the models' \textit{knowledge} by verifying their awareness of different entity readings. To this end, we use the prompt template \textit{``Tell me about <entity-type> <entity>''} (see more details about prompts in Appendix \ref{app:prompts}) to verify that all LLMs generate meaningful output conforming to world knowledge:
i.e., when asked about ``fruit Apple,'' the models provides some information about the fruit, and when asked about ``company Apple,'' it should discuss a company named Apple. 
Next, we ensure that all entities in our experiments pass Study 1, resulting in a final list of 49 entities shown in Table~\ref{tab:categories}.
Additionally, we conduct Study 1a, where we directly query the models about their awareness of ambiguity (\textit{``Can <entity> mean anything other than <entity-type>? Answer only with Yes or No.''}).

\paragraph{Study 2: Eliciting Preferences (K + A).} 
Intuitively, if a model has been exposed to the company \textit{Cisco} far more often than the location \textit{Cisco} (city in Texas), we would assume that it is biased towards the former interpretation. We refer to it as its \textbf{preferred reading}. 
To determine each model's preferred reading (for later  analysis on whether a model's behaviour is affected by it (RQ2), i.e., whether answer correctness increases (decreases) with preferred (alternative) entity interpretation) we prompt LLMs with: \textit{``Group the following entities according to what they all have in common: <entities>''}, where \textit{<entities>} refers to all members of a given category. 
Since the grouping criteria are unspecified (meaning there is no expected \emph{correct} entities interpretation), we assume the model will adopt its preferred reading of the given entity type when forming the grouping.

\begin{figure}[t!]
\vspace{-8pt}
\includegraphics[scale=0.28]{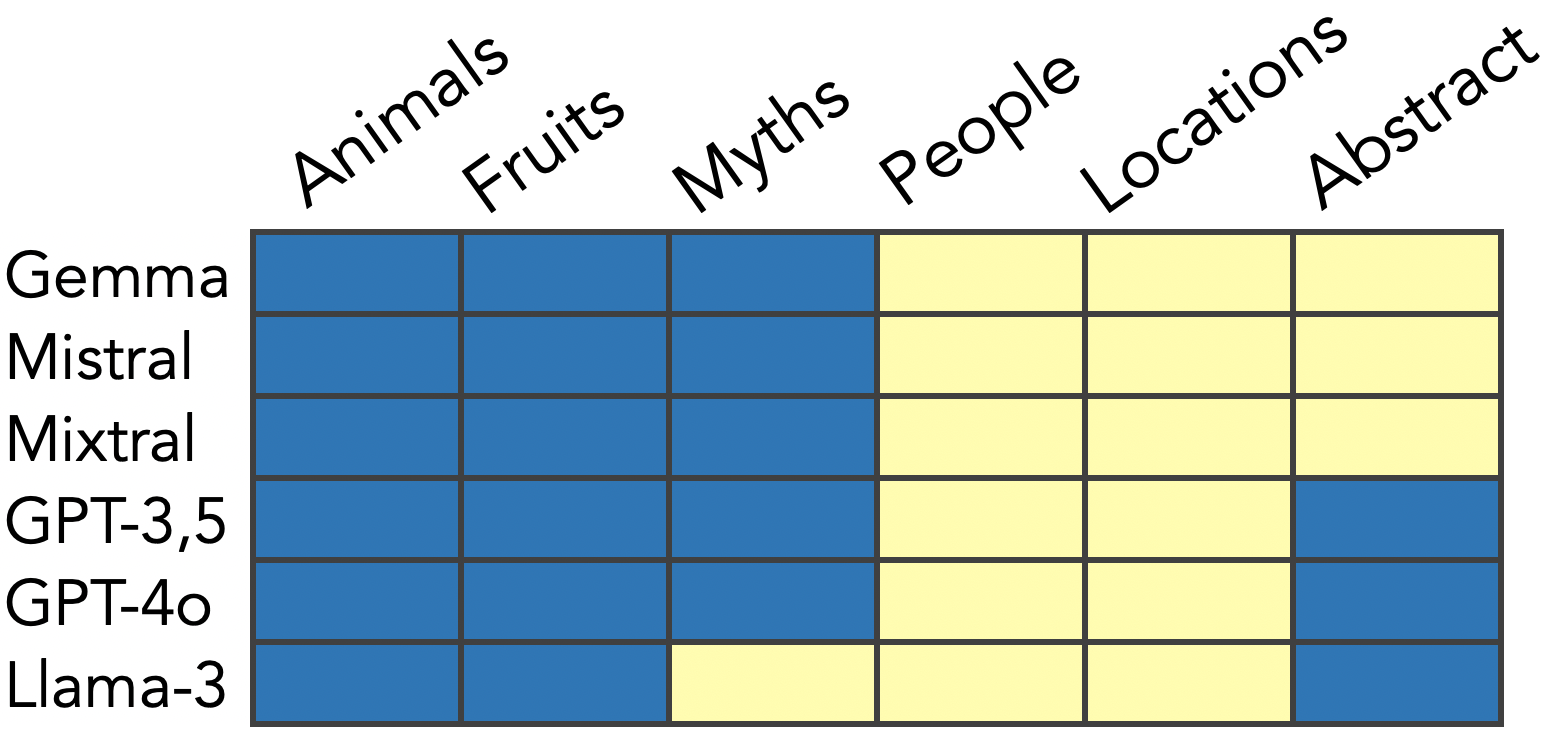}
\vspace{-22pt}
\caption{Models preferred readings discovered in Study~2 (blue for non-company, yellow for company; e.g., all analyzed models prefer the `company' reading for entities from the \textit{people} category).}
\vspace{-4pt}
\label{fig:preferred_reading}
\end{figure}

\paragraph{Study 3: Knowledge to Application (K~$\rightarrow$~A).} 
We proceed to test the \textit{knowledge application} ability by examining if LLMs adopt the correct reading for ambiguous entities 
when answering simple questions related to entity properties. 
Importantly, only the correctness of the chosen reading is evaluated, not the factual accuracy of models' answers.
We use the prompt template \textit{``Provide the <entity-property> for <entity>.''} to evaluate if LLMs are capable to implicitly infer \textit{<entity-type>} (for example, infer company when prompted for founding year). We compare their performance against a non-ambiguous baseline with explicit entity hint, which serves as an upper bound: \textit{``Provide the <entity-property> for <entity-type><entity>.''}

\paragraph{Study 4: Applying to Knowing (A~$\rightarrow$~K).} %
Finally, we aim to assess how consistent the models are with their internal knowledge. 
We automatically retrieve factual information (\textit{<info>}) from the model replies in Study~3 
and prompt the same model back to see if it confirms or denies the correctness of information provided. 
For example, the knowledge about the non-company reading of ``animals'' entities is checked with prompts like \textit{"Does an animal X have <info> speed?"}. %
In such setup, %
we, again, do not verify the factual correctness of the responses, but rather operate under a closed world assumption and focus only on consistency within model's internal knowledge, ensuring fair comparison across models of different sizes.

\begin{table*}[t]
    \centering
    \begin{tabularx}{\textwidth}{|l|c|c|c|c|c|c|X|}
        \hline
        \multirow{2}{*}{Model} & 
        \multicolumn{2}{c|}{Preferred Reading} & 
        \multicolumn{2}{c|}{Alternative Reading} & 
        \multicolumn{3}{c|}{Average} \\
        \cline{2-8} 
        & \footnotesize{\texttt{prop} X} &
        \footnotesize{\texttt{prop type} X} & 
        \footnotesize{\texttt{prop} X}& 
        \footnotesize{\texttt{prop type} X} & 
        \footnotesize{\texttt{prop} X} &
        \footnotesize{\texttt{prop type} X} & 
        \footnotesize{Agg.}\\
        \cline{2-8}
        \hline
        Gemma \footnotesize{\cite{gemmateam2024gemma}} & 87.8 & 95.9 & 63.3 & 69.4 & 75.6 & 82.7 & 77.6 \\
        Mistral \footnotesize{\cite{jiang2023mistral}} & 77.6 & \textbf{100.0} & 63.3 & 87.8 & 70.5 & 93.9 & 82.2 \\
        Mixtral \footnotesize{\cite{jiang2024mixtral}} & 77.6 & \textbf{100.0} & 75.5 & 85.7 & 76.6 & 92.9 & 84.8\\
        GPT-3.5 \footnotesize{\cite{gpt-3.5}} & 87.8 & \textbf{100.0} & 75.5 & 77.6 & 81.7 & 88.8 & 85.3\\
        GPT-4o \footnotesize{\cite{gpt-4o}} & \textbf{93.9} & \textbf{100.0} & 83.7 & 89.8 & \textbf{88.8} & \textbf{94.9} & \textbf{91.9}\\
        Llama-3 \footnotesize{\cite{llama3}} & 87.8 & 98.0 & \textbf{85.7} & \textbf{100.0} & 86.8 & 99.0 & 89.9\\
        \hline
        \textbf{Average} & 85.4 & \textbf{99.0} & 74.5 & 85.1 & 80.0 & 90.5 & 85.3\\
        \hline
    \end{tabularx}
    \vspace{-5pt}
    \caption{Results of Study 3: Knowledge to Application (\% of replies with correctly adopted reading).
    \texttt{``prop''} stands for reading-specific property, \texttt{``type''} - for the corresponding entity type (see Table~\ref{tab:categories}).
    For example, \texttt{prop X} prompt: \textit{``Provide the founding year of Apple''}, \texttt{prop type X} prompt: \textit{``Provide the founding year of \textbf{\textit{company}} Apple.''}}
    \label{tab:understand_apply}
\end{table*}

\section{Results and Discussion}
\label{sec:results}

\paragraph{RQ1: How well can LLMs implicitly disambiguate entity types?} By design, all entities passed Study~1, meaning the LLMs are aware of both entity readings and can provide relevant information about the associated objects or companies.
However, when directly prompted about ambiguity awareness (Study~1a), only Llama-3 recognized ambiguity in all entities, while other models struggled with this task (see Table \ref{tab:study_1_add} in Appendix). 
The highest ambiguity confirmation rates were demonstrated for \textit{Animals} and \textit{Fruits} entities, while for other types, models confirmed ambiguity for no more than 53\% of entities on average\footnote{Model responses in Studies 1-3 were manually reviewed; see Appendix \ref{app:annotation_details} for more details about results evaluation.}.

Figure~\ref{fig:preferred_reading} summarizes the results of Study 2. 
For four of six entity types, all models favored the same interpretation: non-company for \textit{Animals} and \textit{Fruits}, and company for \textit{People} and \textit{Locations}; for the others, the preferred reading varies by model.
Notably, the large Llama-3 and Mixtral models, despite ultimately grouping entities based on a single reading, demonstrate an understanding of their ambiguity
(e.g., Mixtral: \textit{``All of the words you've listed are common names for either a type of animal or a brand [...]''})\footnote{Here and throughout, the example are presented with omissions. For the full outputs and additional findings, please refer to Appendix \ref{app:studies}.}.
Another observation is that all the models tend to provide more varied responses for \textit{Abstract} and \textit{Myths} entities. We hypothise that this is due to their higher ambiguity, which leads to detection of other categories beyond just companies and predefined entity type (e.g., \textit{Mars} and \textit{Vulcan} may be seen neither as companies nor as mythological characters, but as planets, and grouped correspondingly). 
This hypothesis is supported by the entity ambiguity estimations based on Wikipedia disambiguation pages (the average number of entries on disambiguation pages across all entities in the group): $79.9$ for \textit{Abstract} and $69.1$ for \textit{Myth} entities, compared to, for example, $22.3$ for \textit{Locations} (see Table \ref{tab:entity_popularity_summary} in Appendix \ref{app:sec:popularity}).
However, these mixed responses do not clarify whether the model successfully recognizes ambiguity, as including both ``company'' and ``non-company'' reading in the same entity list could indicate either a misunderstanding of entity ambiguity (i.e., the model identifies some entities as companies and others not, despite evidence from Study~1) or a preference for specific readings for certain entities.

The results of Study 3 are presented in Table~\ref{tab:understand_apply}. 
On average, LLMs adopt the correct reading for 85.3\% of entities. However, even with non-ambiguous prompts that include hint so that the entity type is supposed to be clear (e.g., \textit{``Provide the founding year for \underline{company} Apple''}) LLMs reach 90.5\%, thus fail with $\sim$10\% of all entities. 
We also observe a correlation between model size and the amount of incorrect readings: Gemma, the smallest model, has the lowest performance with only an average of $77.6\%$ correctly picked readings, while the large Llama-3 and GPT-4o models perform best at $\sim$90\%. 
Overall, we can conclude that while all models have previously demonstrated knowledge about different readings of ambiguous entities, they struggle to apply this knowledge when choosing the correct reading in their responses.

\paragraph{RQ2: How (much) is the ability to infer the correct entity type biased towards ``preferred readings''?} 
When we break the results further down into preferred and alternative readings, we observe striking differences. For preferred readings, LLMs achieve an average accuracy of 85.4\% with ambiguous prompts, rising to nearly perfect performance (99\%) with non-ambiguous prompts that include hints. However, the results are substantially lower for non-preferred (alternative) readings, where performance drops to 74.5\%/85.1\%. This shows a clear bias across all models toward preferred readings. %
Among the common systematic errors is the model providing company foundation dates when asked for a \textit{date of birth}, even with non-ambiguous prompts: e.g., \textit{“Provide the date of birth for a person Boeing,”}--Mixtral response: \textit{``I'm sorry for any confusion, but Boeing is not a person, it's a multinational corporation [...]''}. 
Moreover, certain entities are consistently assigned incorrect readings by the models, correlating with their popularity, as discussed next.  

\begin{figure*}[t!]
\includegraphics[scale=0.4]{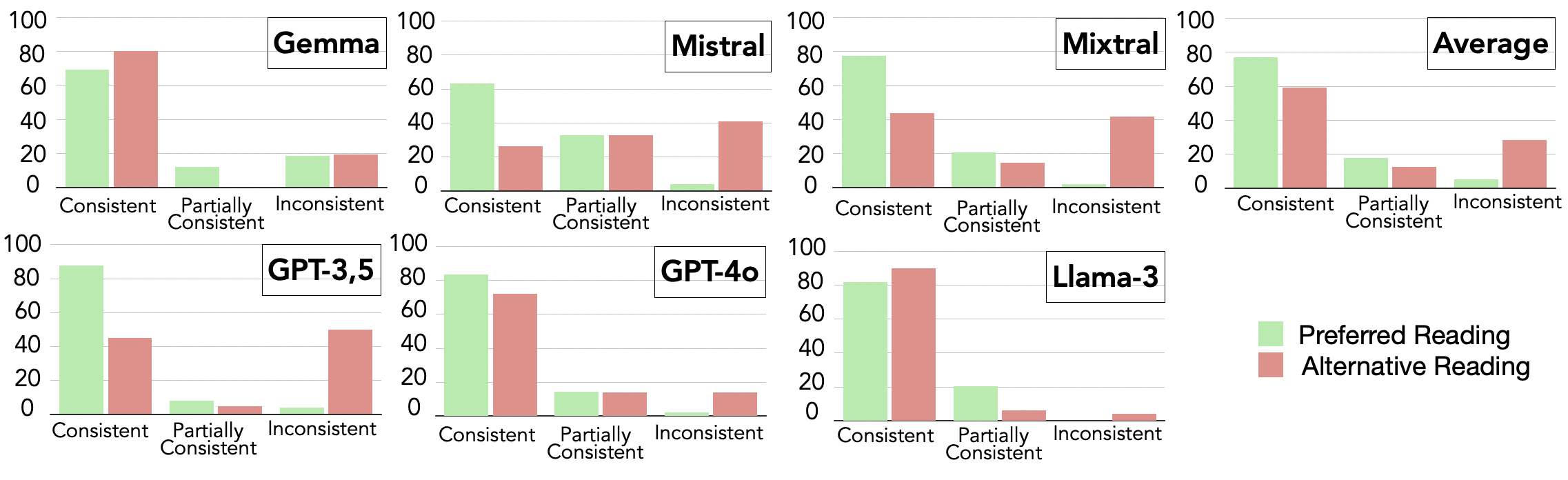}
\vspace{-25pt}
\caption{Results of Study 4 (\% of entities). ``\textit{Consistent}'' entities are those for which the model reaffirmed all provided information in Study 3. ``\textit{Partially Consistent}'' entities are those where some information was reaffirmed but not all, while ``\textit{Inconsistent}'' entities are those for which all previously provided information was denied. The exact numbers are provided in Appendix \ref{app:studies} (Table \ref{app:tab:study_4}).}
\label{fig:appl_know}
\end{figure*}

We hypothesize that a model's preferred reading of particular entities is influenced by their frequencies in the pre-training corpus. For example, if \textit{nike} mostly appears in the context of companies (as opposed to the context of a Greek goddess), we would expect this meaning to dominate over other readings. We follow \citet{mallen2023trust} and use Wikipedia popularity as a proxy for entity type frequency (see Table~\ref{tab:entity_popularity_summary}). 
Our experiments indeed demonstrate the strong correlation between the entity popularity and a preferred reading. 
For example, whenever a model created a separate ``companies'' group while categorizing \textit{Myths} entities, the group consistently included either just \textit{Amazon} or both \textit{Amazon} and \textit{Nike}; e.g., GPT-3.5: \textit{``1. Mythological entities: [...] 2. Companies/brands: Amazon''}, or GPT-4o: \textit{``[...] 1. Brands and Companies:    - Nike - Amazon 2. Greek Mythology:  [...]}. 
This disparity can be attributed to the much higher popularity of these companies: Amazon's Wikipedia page has over 38 million views, Nike's 18 million, while Hermes, the third most popular company in \textit{Myths}, has only 3 million.
As a result, models tend to favor \textit{company} interpretations, regardless of the prompt.
In Study 3, \textit{all} models fail to provide the non-company information about these entities with underspecified prompts. For example,  when prompted with \textit{``Provide the gender for Amazon''}, Mixtral responded with: \textit{``Amazon.com, Inc. is a company, and as such, it does not have a gender. [...]''}.
The errors in providing people birth dates can also be attributed to the significantly higher popularity of the \textit{company} interpretation of a majority of \textit{People} entities. For instance, Boeing's popularity as a \textit{company} is nearly 1.5 times that of its founder, William E. Boeing, leading models to never adopt the \textit{person} interpretation when prompted with \textit{“Provide the date of birth for Boeing.”} Conversely, when the popularity of the \textit{person}-entity exceeds that of the \textit{company}-entity (e.g., Versace, Tesla, Disney), models tend to adopt correct interpretation; e.g., Gemma: \textit{“Tesla was born on July 10, 1856”}, and Llama-3: \textit{“You're referring to Walt Disney! Walter Elias Disney [...] was born on December 5, 1901.”}
These findings clearly indicate that the models exhibit a bias toward one reading of ambiguous entities over another, influenced by the popularity of the entities in those readings.

\paragraph{RQ3: Can LLMs self-verify their answers, given that they successfully disambiguated them?} 
Finally, we investigate whether successful disambiguation implies that the model's answers can be self-verified - in other words, whether a model remains consistent with its previous statements (Study~4).
As Figure \ref{fig:appl_know} shows, \textit{none} of the tested models confirmed all the knowledge provided in the previous study. 
On average, LLMs show higher consistency with preferred readings, with Llama-3 being the most self-consistent (over 80\% across all readings), while Mistral performed the worst (over 30\% of preferred and 60\% of alternative readings unverified).
To analyze the non-confirmation cases, we conducted additional experiments with a modified prompt requesting an explanation (e.g., \textit{``Is December 5 1901 the date of birth of a person Disney? Answer with Yes or No and provide explanation.''}).
Surprisingly, the majority of results were like: (GPT-3.5) \textit{``No. December 5, 1901 is not the date of birth of Walt Disney. Walt Disney was actually born on December 5, 1901.''} 
These findings show that current state-of-the-art LLMs often fail to self-verify their answers, even when they provide the same information immediately afterward.

\section{Conclusion}

We find that state-of-the-art LLMs perform poorly at implicitly disambiguating entity types. Furthermore, their performance is biased by a preferred reading that is heavily influenced by the popularity of the entities. %
Finally, we find that LLMs cannot reliably self-verify their own answers. %
Our results highlight the lack of self-consistency as an open challenge of current LLMs.

\section{Limitations}

In this study, we adopt a very generic definition of ambiguity, distinguishing between company-related and non-company-related company vs. non-company readings across different entity types. 
A more thorough investigation into the degrees of polysemy associated with different entity types should be included in a follow up study. 
Moreover, the properties of the entities might also contain a certain level of ambiguity that we are not thoroughly addressing in this work.

\section{Acknowledgement}

We thank the anonymous reviewers for their helpful suggestions and constructive feedback. This research has been funded by the Vienna Science and Technology Fund (WWTF)[10.47379/VRG19008] “Knowledge-infused Deep Learning for Natural Language Processing” and ERC Consolidator Grant DIALECT 101043235.

\bibliography{anthology,custom}

\appendix

\section{Entity Popularity and Ambiguity}
\label{app:sec:popularity}
\begin{table*}[!htp]
    \centering
    \small
    \begin{tabularx}{\textwidth}{|c|l|r|r|r|r|r|r|r|}
        \cline{1-9}
        \multirow{2}{*}{\textbf{Type}} & 
        \multirow{2}{*}{\textbf{Entity}} &
        \multirow{2}{*}{\textbf{Ambiguity}} &
        \multicolumn{3}{c|}{\textbf{Company Reading}} &
        \multicolumn{3}{c|}{\textbf{Non-Company Reading}} \\ 
        \cline{4-9}
        &&& \textbf{Popularity} & \textbf{\texttt{prop X}} & \textbf{\texttt{prop type X}} & \textbf{Popularity} & \textbf{\texttt{prop X}} & \textbf{\texttt{prop type X}}\\ 
        \cline{1-9}
        
        \multirow{8}{*}{\STAB{\rotatebox[origin=c]{90}{Animal}}}
        & Penguin & 55 & 1,330,112 & 100.0 & 100.0 & 8,965,921 & 100.0 & 100.0 \\
        & Jaguar & 53 & 7,989,902 & 100.0 & 100.0 & 11,939,755 & 0.0 & 100.0 \\
        & Greyhound & 36 & 1,823,476 & 100.0 & 100.0 & 3,380,437 & 33.3 & 100.0 \\
        & Fox & 89 & 3,648,500 & 100.0 & 100.0 & 9,301,784 & 100.0 & 100.0 \\
        & Dove & 50 & 3,796 & 100.0 & 100.0 & 4,244,904 & 50.0 & 83.3 \\
        & Lynx & 78 & 1,057,210 & 100.0 & 100.0 & 6,650,833 & 83.3 & 100.0 \\
        & Puma & 45 & 4,701,402 & 100.0 & 100.0 & 11,554,347 & 83.3 & 100.0 \\
        \cline{2-9}
        & \textbf{Average} & \textbf{58.0} & \textbf{2,936,343} & \textbf{100} & \textbf{100} & \textbf{8,005,426} & \textbf{64} & \textbf{98} \\
        \cline{1-9}
        \multirow{7}{*}{\STAB{\rotatebox[origin=c]{90}{Fruit}}}
        & Apple & 49 & 40,325,969 & 100.0 & 100.0 & 10,948,070 & 33.3 & 100.0 \\
        & Fig & 15 & 129,832 & 100.0 & 83.3 & 2,248,635 & 83.3 & 100.0 \\
        & Mango & 43 & 823,939 & 100.0 & 100.0 & 8,713,110 & 100.0 & 100.0 \\
        & Kiwi & 36 & 293,874 & 100.0 & 100.0 & 6,245,271 & 100.0 & 100.0 \\
        & Papaya & 12 & - & 100.0 & 100.0 & 4,770,845 & 100.0 & 100.0 \\
        & Orange & 103 & 2,007,461 & 100.0 & 100.0 & 7,409,145 & 66.7 & 83.3 \\
        \cline{2-9}
        & \textbf{Average} & \textbf{43.0} & \textbf{8,716,215} & \textbf{100.0} & \textbf{97.2} & \textbf{6,722,513} & \textbf{80.6} & \textbf{97.2} \\
        \cline{1-9}
        \multirow{9}{*}{\STAB{\rotatebox[origin=c]{90}{Myth. Character}}}
        & Pegasus & 86 & 1,773,226 & 33.3 & 83.3 & 4,853,706 & 100.0 & 100.0 \\
        & Vulcan & 79 & 635,380 & 66.7 & 100.0 & 2,673,387 & 0.0 & 100.0 \\
        & Midas & 38 & 187,394 & 83.3 & 83.3 & 3,687,467 & 100.0 & 100.0 \\
        & Nike & 34 & 18,187,528 & 100.0 & 100.0 & 4,375,918 & 33.3 & 100.0 \\
        & Mars & 134 & 259,189 & 33.3 & 100.0 & 19,365,488 & 66.7 & 100.0 \\
        & Hyperion & 62 & 58,794 & 66.7 & 100.0 & 1,316,548 & 83.3 & 100.0 \\
        & Hermes & 56 & 3,426,101 & 83.3 & 100.0 & 10,337,899 & 100.0 & 100.0 \\
        & Amazon & 64 & 38,684,687 & 100.0 & 100.0 & 5,119,820 & 16.7 & 100.0 \\
        \cline{2-9}
        & \textbf{Average} & \textbf{69.1} & \textbf{7,901,537} & \textbf{70.8} & \textbf{95.8} & \textbf{6,466,279} & \textbf{62.5} & \textbf{100.0} \\
        \cline{1-9}
        \multirow{11}{*}{\STAB{\rotatebox[origin=c]{90}{Person}}}
        & Versace & 13 & 7,095,079 & 100.0 & 100.0 & 22,180,811 & 100.0 & 66.7 \\
        & Boeing & - & 10,754,848 & 100.0 & 100.0 & 681,877 & 0.0 & 33.3 \\
        & Ford & 104 & 14,643,256 & 100.0 & 100.0 & 13,966,210 & 83.3 & 50.0 \\
        & Philips & 6 & 5,948,052 & 100.0 & 100.0 & 331,229 & 16.7 & 33.3 \\
        & Levi Strauss & 13 & 3,744,382 & 100.0 & 100.0 & 2,320,188 & 100.0 & 100.0 \\
        & Ferrero & 4 & 3,447,282 & 100.0 & 100.0 & 409,662 & 66.7 & 66.7 \\
        & Tesla & 21 & 23,462,104 & 100.0 & 100.0 & 37,395,340 & 83.3 & 83.3 \\
        & Disney & 58 & 20,938,263 & 100.0 & 100.0 & 31,693,370 & 100.0 & 50.0 \\
        & Dell & 22 & 7,310,499 & 100.0 & 100.0 & 3,558,086 & 16.7 & 33.3 \\
        & Benetton & 5 & 1,864,193 & 100.0 & 100.0 & 378,208 & 50.0 & 50.0 \\
        \cline{2-9}
        & \textbf{Average} & \textbf{27.3} & \textbf{9,920,796} & \textbf{100.0} & \textbf{100.0} & \textbf{11,291,498} & \textbf{61.7} & \textbf{56.7} \\
        \cline{1-9}
        \multirow{9}{*}{\STAB{\rotatebox[origin=c]{90}{Location}}}
        & Cisco & 26 & 1,738,862 & 100.0 & 100.0 & - & 0.0 & 100.0 \\
        & Prosper & 10 & 276,714 & 100.0 & 83.3 & 419,461 & 33.3 & 100.0 \\
        & Patagonia & 12 & 1,055,737 & 100.0 & 100.0 & 11,426,844 & 100.0 & 100.0 \\
        & Montblanc & 5 & 1,306,077 & 100.0 & 100.0 & 5,671,509 & 100.0 & 100.0 \\
        & Amazon & 64 & 38,684,687 & 100.0 & 100.0 & 6,509,535 & 33.3 & 100.0 \\
        & Nokia & 13 & 11,446,036 & 100.0 & 100.0 & 332,572 & 0.0 & 83.3 \\
        & Hershey & 24 & 3,929,199 & 100.0 & 100.0 & 1,419,873 & 100.0 & 100.0 \\
        & Eagle Creek & 24 & 55,717 & 100.0 & 100.0 & 2,248 & 83.3 & 100.0 \\
        \cline{2-9}
        & \textbf{Average} & \textbf{22.3} & \textbf{7,311,629} & \textbf{100.0} & \textbf{97.9} & \textbf{3,683,149} & \textbf{58.3} & \textbf{95.8} \\
        \cline{1-9}
        \multirow{11}{*}{\STAB{\rotatebox[origin=c]{90}{Abstract}}}
        & Harmony & 119 & 143,865 & 83.3 & 83.3 & 1,847,278 & 100.0 & 100.0 \\
        & Fidelity & 29 & 3,648,171 & 100.0 & 100.0 & 633,474 & 100.0 & 100.0 \\
        & Allure & 17 & 832,160 & 100.0 & 100.0 & 728,597 & 50.0 & 100.0 \\
        & Vision & 102 & 29,660 & 100.0 & 100.0 & 1,810,577 & 100.0 & 100.0 \\
        & Genesis & 141 & 2,809,401 & 50.0 & 100.0 & 6,338,641 & 100.0 & 100.0 \\
        & Tempo & 59 & 27,507 & 100.0 & 100.0 & 5,416,890 & 66.7 & 100.0 \\
        & Triumph & 45 & 351,267 & 100.0 & 100.0 & 1,132,962 & 83.3 & 100.0 \\
        & Vanguard & 128 & 6,661,130 & 100.0 & 100.0 & 1,059,408 & 16.7 & 83.3 \\
        & Pioneer & 95 & 1,058,945 & 100.0 & 100.0 & 521,227 & 66.7 & 100.0 \\
        & Zenith & 64 & 753,374 & 100.0 & 100.0 & 1,602,303 & 100.0 & 100.0 \\
        \cline{2-9}
        & \textbf{Average} & \textbf{79.9} & \textbf{1,631,548} & \textbf{93.3} & \textbf{98.3} & \textbf{2,109,136} & \textbf{78.3} & \textbf{98.3} \\
        \cline{1-9}
    \end{tabularx}
    \caption{Summary of entity types their characteristics:  ambiguity and popularity. Following \citet{mallen2023trust}, we evaluate the popularity and ambiguity of each entity based on Wikipedia page views and the number of pages references to on the Wikipedia entity disambiguation page, respectively.
    Data collected on: June 10, 2024.
    Dash means there is no Wikipedia disambiguation page for this entity.
    Additionally, we provide the model performance on each entity demonstrated in Study 3, aggregated across the models.}
    \label{tab:entity_popularity_summary}
\end{table*}

The entity selection process was based on the authors’ intuition, followed by filtering the initial list to include only entities that all LLMs recognize in both their main and company readings (that is, the entities have passed Study 1). 

\begin{figure*}[t!]
\centering
\includegraphics[scale=0.17]{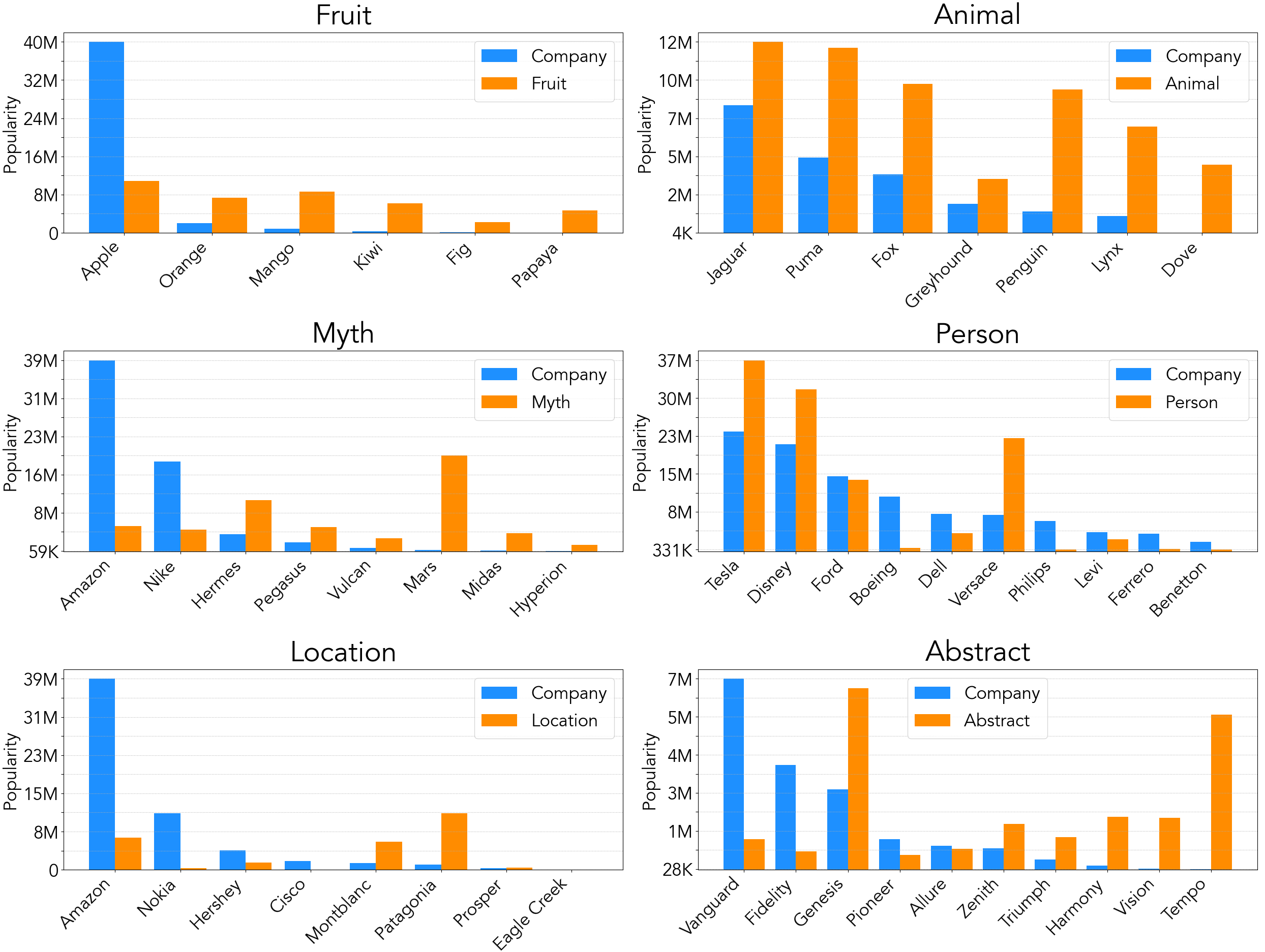}
\vspace{-8pt}
\caption{Popularity distribution of company and non-company readings of all 49 entities involved in our studies.}
\vspace{-10pt}
\label{app:fig:entities}
\end{figure*}

In Table \ref{tab:entity_popularity_summary}, we present additional information about all entities utilized in our experiments. Following \citet{mallen2023trust}, we assess the popularity of each entity (in our context: each entity's interpretation, such as company-related and non-company-related) based on Wikipedia page views over the past nine years. In instances of ambiguity within a single interpretation (e.g., multiple companies sharing the same name, or multiple individuals with the same surname), we selected the most popular one. Furthermore, we estimated the ambiguity of each entity using its corresponding Wikipedia disambigation page, for example: \texttt{https://en.wikipedia.org/wiki/
Jaguar\_(disambiguation)}. Specifically, we counted the number of pages listed on the disambiguation page, providing a preliminary estimate of the number of real-world entities to which the term could refer.

Additionally, in order to evaluate correlation between the performance of the models on individual entities and the popularity of these entities, we aggregated the results of Study 3 across all models for each entity. 
Specifically, for each of the entity readings, we counted how many times each model selected a correct interpretation when providing response to a relevant prompt and calculated the average. 
For example, the performance of the models for entity \textit{Jaguar} in its company reading was aggregated from the replies of all models to the prompt \textit{``Provide the founding year for the company Jaguar''}.
The plots illustrating the popularity of the entities are shown in Figure \ref{app:fig:entities}.

\section{Annotation Details}
\label{app:annotation_details}

To evaluate the LLMs' output, we opted for a manual annotation of all models generations.
To keep the annotation process as simple and clear as possible as well as preserve experiments scalability, we designed it to resemble automatic evaluation rather than typical human annotation, ensuring that it remains objective and unambiguous.
For example, determining whether a model's response about \textit{apple} refers to the company or the fruit is straightforward by examining the model's reply. 
\bigskip

\noindent{}Prompt: \textit{``Provide the color for Apple''}, Llama-3: \textit{``The iconic color often associated with Apple is Silver/Space Gray, but if you're thinking of a more specific brand element, it's likely the bright, bold, and vibrant color: **Red**''} (obviously, this response refers to the entity \textit{apple} as a company).
\bigskip

\noindent{}Prompt: \textit{``Provide the color for a fruit Apple''}, Llama-3: \textit{``The typical color of an Apple is RED, but apples can also come in other colors such as: * Green (Granny Smith, Green Apple)* Yellow (Golden Delicious, Yellow Apple)* Red with Yellow or Green stripes (Gala, Fuji)* Purple (some heirloom varieties) But, if I had to pick one, I'd say the most iconic color for an apple is RED!''} (this response addresses the color of \textit{apple} as a fruit).
\bigskip

The same is valid for the annotation in Study 3:  
\bigskip

\noindent{}Prompt: \textit{``Provide the date of birth for Boeing''}, Llama-3: \textit{``Boeing’s founding date is July 15, 1916, when William E. Boeing and Conrad Westervelt founded [...], which later became The Boeing Company.''} (the annotation is clear: the model adopted the ``company'' reading. Since we are not assessing factual correctness, the accuracy of dates is irrelevant, and we do not have gold labels for it).

\begin{table*}[ht!]
    \centering
    \begin{tabularx}{\textwidth}{|Y|YYYYYY|Y|}
    \hline
    Model & Animals & Fruits & Myths & People & Locations & Abstract & \textbf{Average} \\ \hline
    Gemma & \textbf{100.0} & \textbf{100.0} & 37.5 & 0.0 & 12.5 & 10.0 & 43.3 \\
    Mistral & \textbf{100.0} & 83.8 & 75.0 & 10.0 & 75.0 & 90.0 & 72.3 \\
    Mixtral & 71.4 & 50.0 & 0.0 & 0.0 & 30.0 & 50.0 & 93.1 \\
    GPT-3.5 & 57.1 & \textbf{100.0} & 0.0 & 10.0 & 12.5 & 10.0 & 31.6 \\
    GPT-4o & \textbf{100.0} & \textbf{100.0} & \textbf{100.0} & 60.0 & \textbf{100.0} & 90.0 & 91.7 \\
    LLaMa-3 & \textbf{100.0} & \textbf{100.0} & \textbf{100.0} & \textbf{100.0} & \textbf{100.0} & \textbf{100.0} & \textbf{100.0} \\ \hline
    \textbf{Average} & 80.1 & \textbf{89.0} & 52.1 & 50.0 & 52.8 & 50.0 & 72.0 \\ \hline
    \end{tabularx}
    \caption{The results of experiments with direct prompting the model about the ambiguity (\textit{"Can <entity> mean anything else but <entity-type>? Answer only with Yes or No."}).}
    \label{tab:study_1_add}
\end{table*}

\section{Prompts}
\label{app:prompts}
The full lists of prompts used in our experiments are to be found in Table \ref{tab:all_prompts_study}.
At a high level, all our studies can be viewed as different variants of the same prompt, designed to elicit the same information (entity types), and extensive prompt engineering is beyond the scope of this work.
However, 
we conducted some experiments on prompt variation, considering two types of modification: one at the formatting level and the other at the semantic level.

\paragraph{Formatting level.}

To assess the variability in model performance across different prompt formats, we utilized the tool \texttt{FormatSpread} \cite{sclar2023quantifying}. 
This tool was initially designed to generate semantically equivalent variations of a given prompt to identify the most effective format.
For our study, we randomly selected 20 prompt formats from those generated by \texttt{FormatSpread}, applied these formats to all our original prompts, and repeated the same experimental and evaluation procedures.
Examples of these formats include: (original) \textit{Tell me about a company called Amazon}, (1) \textit{Tell me about a company called - Amazon}, (2) \textit{TELL ME ABOUT A COMPANY CALLED::: Amazon}, (3) \textit{TELL ME ABOUT A COMPANY CALLED\string\nAmazon}, (4) \textit{Tell Me About A Company Called\string\tAmazon}, (5) \textit{tell me about a company called Amazon}, (6) \textit{TELL ME ABOUT A COMPANY CALLED : Amazon}, etc. 
However, we did not observe any significant differences in the results compared to those obtained with the original prompts.

\paragraph{Semantic level.}
At the semantic level, we tested different paraphrasings to select the prompts for the final experimental setup. A notable difference between paraphrasings was observed only in Study 2, where we used four variants of the initial prompt (see Table \ref{tab:all_prompts_study}, Study 2) and aggregated the model responses through majority voting.

\section{Results: Further Discussion}
\label{app:studies}

\paragraph{Study 1.}
The results of directly prompting the model to determine its awareness of ambiguity, using the prompt \textit{``Can <entity> mean anything else but <entity-type>? Answer only with Yes or No.''}, are provided in Table \ref{tab:study_1_add}.
As it becomes clear, despite possessing knowledge about the different meanings of each entity (as proven by Study~1, see Section \ref{sec:results}), the models often fail to provide this information when asked directly. For example, Mixtral often denies other interpretations, justifying this by claiming that there is one clear meaning of the entity, although it can be used for other purposes (for example, \textit{``No, Eagle Creek cannot mean anything else in general usage. It is primarily a geographic location, specifically a creek name occurring in various places in the United States. However, like many place names, it can be used as a proper noun in other contexts, such as brand names (e.g., Eagle Creek luggage).''}).
From this observation, we make two assumptions: (1) each model may have a preferred interpretation for each entity and entity type, an hypothesis we intend to explore in Study~2, and (2) a more carefully considered experimental setup is required, rather than straightforwardly querying the model about ambiguity, which was one of the motivations behind the grouping task approach we adopted for  Study 2. 

\paragraph{Study 2.}

Complete examples of model responses for \textit{Myths} entities are as follows:
\bigskip

\noindent{}Prompt: \textit{``Group the following entities: Amazon, Mars, Vulcan, Hermes, Hyperion, Nike, Midas, Pegasus.''}, GPT-3.5: \textit{``1. Mythological entities: Mars, Vulcan, Hermes, Hyperion, Nike, Midas, Pegasus. \textbf{2. Companies/brands: Amazon}''}, GPT-4o: \textit{``Certainly! These entities can be grouped based on their categories or mythological references: By category: \textbf{1. Brands and Companies:    - Nike    - Amazon} 2. Greek Mythology:    - Nike (Also a brand, but originally the Greek goddess of victory)    - Pegasus    - Hermes 3. Roman Mythology:    - Mars    - Vulcan 4. Mythological figures and references from various mythologies:    - Hyperion (Greek mythology, a Titan)    - Midas (Greek mythology, king with the golden touch) By mythology: 1. Greek Mythology:    - Nike    - Pegasus    - Hermes    - Hyperion    - Midas 2. Roman Mythology:    - Mars (Roman counterpart to the Greek god Ares)    - Vulcan (Roman counterpart to the Greek god Hephaestus) \textbf{3. Modern entities named after mythological references: - Nike} (brand named after the Greek goddess of victory) \textbf{- Amazon} (company name inspired by the Amazon warriors of Greek mythology)''}, Mixtral: \textit{``Sure, I will group these entities into the following categories: 1. Planets: Mars, 2. Mythological figures: Hyperion, Midas, Nike, Hermes, Vulcan, Pegasus, \textbf{3. Companies: Amazon.} Explanation: 1. Mars is a planet in our solar system. 2. Hyperion, Midas, Nike, Hermes, Vulcan, and Pegasus are all figures from Greek and Roman mythology. Hyperion is a Titan and the father of the Sun, Moon, and Dawn. Midas was a king who was granted the power to turn everything he touched into gold. Nike is the goddess of victory, Hermes is the god of trade, thieves, travelers, and athletes, Vulcan is the god of fire including the fire of volcanoes, and Pegasus is a winged horse. 3. Amazon is a multinational technology company that focuses on e-commerce, cloud computing, digital streaming, and artificial intelligence. It is one of the world's largest online marketplaces, AI assistant providers, and cloud computing platforms.''}

\begin{table*}[ht!]
    \centering
    \small 
    \begin{tabularx}{\textwidth}{|l|*4{>{\centering\arraybackslash}r|}*4{>{\centering\arraybackslash}r|}}
        \cline{1-9}
        \multirow{3}{*}{\textbf{Model}} & \multicolumn{4}{c|}{\textbf{Preferred Reading}} & \multicolumn{4}{c|}{\textbf{Alternative Reading}} \\ \cline{2-9}
         & Consistent & %
         \begin{tabular}{@{}c@{}}Partially \\ Consistent\end{tabular} &
         Inconsistent & Sum & Consistent & %
         \begin{tabular}{@{}c@{}}Partially \\ Consistent\end{tabular} & Inconsistent & Sum \\ \cline{2-9}

        & \multicolumn{8}{c|}{\textbf{Companies Reading}} \\ \cline{1-9}

        Gemma & 18 (38.3\%) & 1 (2.1\%) & 9 (19.1\%) & 28 & 13 (27.7\%) & 0 (00.0\%) & 6 (12.8\%) & 19 \\ 
        Mistral & 21 (42.9\%) & 7 (14.3\%) & 0 (00.0\%) & 28 & 10 (20.4\%) & 8 (16.3\%) & 3 (6.1\%) & 21 \\
        MiXtral & 24 (50.0\%) & 3 (6.2\%) & 1 (2.1\%) & 28 & 12 (25.0\%) & 5 (10.4\%) & 3 (6.2\%) & 20 \\
        GPT-3.5 & 16 (32.7\%) & 0 (0.0\%) & 2 (4.1\%) & 18 & 19 (38.8\%) & 2 (4.1\%) & 10 (20.4\%) & 31 \\
        GPT-4o & 18 (38.3\%) & 0 (0.0\%) & 0 (0.0\%) & 18 & 24 (51.1\%) & 4 (8.5\%) & 1 (2.1\%) & 29\\
        LLaMa-3 & 24 (49.0\%) & 2 (4.1\%) & 0 (0.0\%) & 26 & 22 (44.9\%) & 1 (2.0\%) & 0 (0.0\%) & 23
        \\ \cline{1-9}

        & \multicolumn{8}{c|}{\textbf{Animals Reading}} \\ \hline
        Gemma & 3 (42.9\%) & 4 (57.1\%) & 0 (0.0\%) & 7 & - & - & - & - \\ 
        Mistral & 2 (28.6\%) & 3 (42.9\%) & 2 (28.6\%) & 7 & - & - & - & - \\
        Mixtral & 6 (85.7\%) & 1 (14.3\%) & 0 (0.0\%) & 7 & - &- & - & -\\
        GPT-3.5 & 5 (71.4\%) & 2 (28.6\%) & 0 (0.0\%) & 7 & - &- & - & -\\
        GPT-4o & 6 (85.7\%) & 0 (00.0\%) & 1 (14.3\%) & 7 & - & - & - & -\\
        LLaMa-3 & 4 (57.1\%) & 3 (42.9\%) & 0 (0.0\%) & 7 & - &- & - & -\\ \cline{1-9}

        & \multicolumn{8}{c|}{\textbf{Fruits Reading}} \\ \cline{1-9}
        Gemma & 5 (83.3\%) & 1 (16.7\%) & 0 (0.0\%) & 6 & - &- & - & -\\
        Mistral & 0 (0.0\%) & 6 (100.0\%) & 0 (0.0\%) & 6 & - &- & - & -\\
        Mixtral & 1 (16.7\%) & 5 (83.3\%) & 0 (0.0\%) & 6 & - &- & - & -\\
        GPT-3.5 & 4 (66.7\%) & 2 (33.3\%) & 0 (0.0\%) & 6 & - &- & - & -\\
        GPT-4o & 5 (83.3\%) & 1 (16.7\%) & 0 (0.0\%) & 6 & - &- & - & -\\
        LLaMa-3 & 2 (33.3\%) & 4 (66.7\%) & 0 (0.0\%) & 6 & - &- & - & -\\ \cline{1-9}
        
        & \multicolumn{8}{c|}{\textbf{Myths Reading}} \\ \hline
        Gemma & 8 (100.0\%) & 0 (0.0\%) & 0 (0.0\%) & 8 & - &- & - & -\\
        Mistral & 8 (100.0\%) & 0 (0.0\%) & 0 (0.0\%) & 8 & - &- & - & -\\
        Mixtral & 7 (87.5\%) & 1 (12.5\%) & 0 (0.0\%) & 8 & - &- & - & -\\
        GPT-3.5 & 8 (100.0\%) & 0 (0.0\%) & 0 (0.0\%) & 8 & - &- & - & -\\
        GPT-4o & 8 (100.0\%) & 0 (0.0\%) & 0 (0.0\%) & 8 & - &- & - & -\\
        LLaMa-3& - &- & - & - & 8 (100.0\%) & 0 (0.0\%) & 0 (0.0\%) & 8 \\ \cline{1-9}
        
        & \multicolumn{8}{c|}{\textbf{People Reading}} \\ \cline{1-9}
        Gemma & - & - &- & - & 5 (71.4\%) & 0 (0.0\%) & 2 (28.6\%) & 7\\
        Mistral & -& - &- & - & 0 (0.0\%) & 0 (0.0\%) & 10 (100.0\%) & 10\\
        Mixtral & -& - &- & - & 1 (10.0\%) & 0 (0.0\%) & 9 (90.0\%) & 10\\
        GPT-3.5 & -& - &- & - & 0 (0.0\%) & 0 (0.0\%) & 6 (100.0\%) & 6\\
        GPT-4o & - & - &- & - & 4 (50.0\%) & 1 (12.5\%) & 3 (37.5\%) & 8\\
        LLaMa-3 & - & - &- & - & 6 (60.0\%) & 2 (20.0\%) & 2 (20.0\%) & 10\\ \cline{1-9}
        
        & \multicolumn{8}{c|}{\textbf{Locations Reading}} \\ \cline{1-9}
        Gemma& -& - &- & - & 5 (100.0\%) & 0 (0.0\%) & 0 (0.0\%) & 5\\
        Mistral& -& - &- & - & 0 (0.0\%) & 3 (37.5\%) & 5 (62.5\%) & 8\\
        Mixtral& -& - &- & - & 0 (0.0\%) & 0 (0.0\%) & 8 (100.0\%) & 8\\
        GPT-3.5& -& - &- & - & 0 (0.0\%) & 0 (0.0\%) & 5 (100.0\%) & 5\\
        GPT-4o& -& - &- & - & 3 (50.0\%) & 1 (16.7\%) & 2 (33.3\%) & 6\\
        LLaMa-3& -& - &- & - & 8 (100.0\%) & 0 (0.0\%) & 0 (0.0\%) & 8\\ \cline{1-9}
        
        & \multicolumn{8}{c|}{\textbf{Abstract Reading}} \\ \cline{1-9}
        Gemma & - & - &- & - & 10 (100.0\%) & 0 (0.0\%) & 0 (0.0\%) & 10 \\
        Mistral & - & - &- & - & 3 (30.0\%) & 5 (50.0\%) & 2 (20.0\%) & 10 \\
        Mixtral & - & - &- & - & 8 (80.0\%) & 2 (20.0\%) & 0 (0.0\%) & 10 \\
        GPT-3.5 & 10 (100.0\%) & 0 (0.0\%) & 0 (0.0\%) & 10 & - &- & - & - \\
        GPT-4o & 4 (40.0\%) & 6 (60.0\%) & 0 (0.0\%) & 10 & - &- & - & -\\
        LLaMa-3 & 10 (100.0\%) & 0 (0.0\%) & 0 (0.0\%) & 10 & - &- & - & -\\ \cline{1-9}

    \end{tabularx}
    \caption{Results from Study 4 (absolute numbers and the percentange of all replies; the graphical interpretation of these findings is illustrated in Figure \ref{fig:appl_know}). For non-company readings, we present the results for each group separately. Depending on whether a model prefers this reading or not, the results are provided in the corresponding column. Additionally, please note that the percentages are calculated based on the number of entities for which the corresponding model provided information in Study 3, not the total number of entities used in the experiments. For example, the Gemma model has provided information about the location for only 5 entities out of 8; therefore, the percentage of each category (consistent, partially consistent, and inconsistent) would be calculated based on 5, not 8 (in contrast to, for instance, the Mistral model which provided information for all 8 entities).
    \newline\newline\newline\newline}
    \label{app:tab:study_4}
\end{table*}

\paragraph{Study 3.}
As discussed in the paper, models more successfully recognize and provides information for entities in their preferred reading. For example, all models successfully provided the required information about \textit{Animals} and \textit{Fruits} in their non-company readings (e.g., speed and color, respectively)--see Table \ref{app:tab:study_4} values in the first \textit{Sum} column. In contrast, several models failed to provide information about the date of birth and geographic locations for some of the \textit{People} and \textit{Locations} entities (the second \textit{Sum} column in Table \ref{app:tab:study_4}). For instance, only LLaMa-3, Mixtral, and Mistral models managed to provide reasonable answers for all entities related to \textit{Locations}. Other models did not recognize either 2 (GPT-4o) or 3 (Gemma and GPT-3.5) locations from the list, even when given the most detailed prompt:
\bigskip

\noindent{}Prompt: \textit{``Provide the area in square meters for a geographic location of Cisco''}, Gemma: \textit{``I am unable to access real-time information or location-specific data, and am unable to provide information regarding the area of Cisco in square meters. For the most up-to-date and accurate information, please refer to reliable sources such as public maps or official Cisco publications.''}, GPT-4o: \textit{``Cisco is a large multinational company that has numerous office locations and various facilities around the world, so it's challenging to specify the exact area in square meters for a specific "Cisco" location without more information. If you are referring to a particular office or campus, please provide additional details so I can help you more accurately. Otherwise, for the company's worldwide presence, there isn't a single, cohesive geographic area that can be quantified.''}, GPT-3.5: \textit{``I'm sorry, but I cannot provide the exact area in square meters for the geographic location of Cisco as it's a private company and not a specific geographical region. However, if you have a specific location in mind related to Cisco, please provide more information so that I can assist you with finding the area of that specific location.''}
\bigskip

Complete examples of model responses for \textit{People} entities are as follows:
\bigskip

\noindent{}Prompt: \textit{``Provide the date of birth for a person Boeing''}, Mixtral: \textit{``I'm sorry for any confusion, but Boeing is not a person, it's a multinational corporation that designs, manufactures, and sells airplanes, rotorcraft, rockets, satellites, telecommunications equipment, and missiles worldwide. The company was founded on July 15, 1916, but I cannot provide a date of birth for a corporation.''} (labeled as incorrect, regardless of whether the provided year is the correct year of the company's founding), Gemma: \textit{``I am unable to access personally identifiable information or sensitive date of birth information of individuals, including Boeing''} (labeled as correct, as the model adopted the correct ``person'' reading, even though no specific information about this person’s date of birth was provided).
\bigskip

Sometimes the models even mention the name of the founder, yet still adhere to the \textit{company} interpretation when providing the response:
\bigskip

\noindent{}Prompt: \textit{``Provide the date of birth for Boeing.''}, Llama-3: \textit{``Boeing's founding date is July 15, 1916, when William E. Boeing and Conrad Westervelt founded Pacific Aero Products Co., which later became The Boeing Company.''}
\bigskip

In the meantime, the fact that all the models that were included in the experiments indeed possess this knowledge was confirmed by an additional sanity check using completely unambiguous prompts, where the entity X is presented with both first and last names, indicating a preferred \textit{people} interpretation in the most unambiguous way: 
\bigskip

\noindent{}Prompt: \textit{``What is the date of birth of William E. Boeing?''}, Llama-3 \& GPT-3.5: \textit{``William E. Boeing was born on October 1, 1881.''}

\bigskip

Another notable trend is that for \textit{people} entities in their non-company reading, all models perform better with underspecified prompts (\textit{``Provide the date of birth for X''}) and often decline to answer more detailed prompts (\textit{``Provide the date of birth for a person X''}), referring to the absence of ``sensitive'' information about specific individuals.
\bigskip

\noindent{}Prompt: \textit{``Provide the date of birth for Versace''}, GPT-3.5: \textit{``Versace was born on December 2, 1946.''}, Gemma: \textit{``The date of birth for Gianni Versace is October 2, 1955''}.
\bigskip

\noindent{}Prompt: \textit{``Provide the date of birth for a person Versace''}, GPT-3.5 Response: \textit{``I'm sorry, but I am unable to provide the date of birth for a person named Versace as it is not specified. If you have a specific individual in mind, please provide more information so I can assist you further.''}, Gemma Response: \textit{``I am unable to access personally identifiable information or sensitive date of birth information of individuals. Sharing such data would violate privacy laws and ethical guidelines''}.

\paragraph{Study 4.}

In Study 4, we prompted the model about the information it provided in Study 3.
For example, if in Study 3 the Mixtral model responded to \textit{``Provide the area in square meters for a geographic location of Nokia''} with \textit{``...The city of Nokia has a total area of approximately 327.74 sq.km. (or 32,774,000 sq.m.)...''}, in Study 4, it will be prompted with \textit{``Is there a geographic location Nokia with an approximate area of 32,774,000 sq.m.? Answer ONLY with Yes or No. If you cannot answer this question, answer No.''}. 
If the model answers ``Yes'', the knowledge about this entity is reaffirmed. 
If not all knowledge is reaffirmed—e.g., in Study 3, the model provided varying colors for a fruit entity, and in Study 4, when asked to confirm each color separately, it responded ``Yes'' to some and ``No'' to others—the entity is classified as \textit{Partially Consistent}.
If the model replies ``No'' about all information it previously provided about the entity in the group, the entity is counted as \textit{Inconsistent}.
Preliminary experiments revealed that closed-source LLMs yielded inconsistent results across multiple runs of the same prompt; therefore, we conducted five trials per prompt and considered the knowledge confirmed if it was confirmed in at least one run. 
The results are provided in Table \ref{app:tab:study_4}.

The highest ratio of denying the correctness of provided information was observed in the \textit{People} group, where the models were asked to reconfirm the date of birth using the prompt: 
\textit{Is <retrieved-info> the date of birth of a person <entity>? Answer ONLY with Yes or No. If you cannot answer this question, answer No.} 
For instance, GPT-3.5, which was among the top three best-performing models in Study 3, did not reconfirm any of the dates (i.e., it replied with a single token ``No'' to all the requests). 
To clarify such unexpected model behavior, we additionally ran experiments with a modified prompt (\textit{Is <retrieved-info> the date of birth of a person <entity>? Answer with Yes or No and provide an explanation.}) aimed at obtaining some explanation\footnote{Importantly, these types of prompts were used only for analysis. For consistency, we used the same prompts provided in Table \ref{tab:all_prompts_study}, Study 4, for the final results.}. 
We obtained surprising results, such as:

\bigskip

\noindent{}Prompt: \textit{``Is December 5 1901 the date of birth of a person Disney? Answer with Yes or No and provide explanation.'', GPT-3.5: \textit{``No. December 5, 1901 is not the date of birth of Walt Disney. Walt Disney was actually born on December 5, 1901.''}}
\bigskip

\noindent{}Prompt: \textit{``Is February 26 1829 the date of birth of a person Levi Strauss? Answer with Yes or No and provide explanation.'', GPT-3.5: \textit{``No. Levi Strauss, the founder of the famous clothing company Levi Strauss \& Co., was born on February 26, 1829.''`}}
\bigskip

\noindent{}Prompt: \textit{``Is April 26 1925 the date of birth of a person Ferrero? Answer with Yes or No and provide explanation.''}, GPT-4o: \textit{``No, April 26, 1925, is not the date of birth of a person named Ferrero. Michele Ferrero, the Italian entrepreneur known for creating Nutella and other popular confectionery products, was born on April 26, 1925.''}
\bigskip

Thus, in fact, the model does confirm the knowledge, but since its first one-word-response is ``No,'' our evaluation system considers this answer as negative, which we believe is a correct evaluation.
The same confusion occasionally occurs with the \textit{Animal} entities:
\bigskip

\noindent{}Prompt: \textit{``Does an animal Dove have 24-35 miles per hour (38-56 kilometers per hour) speed? Answer ONLY with Yes or No. Answer based on your internal knowledge. Do not include any additional clarification. If you cannot answer this question, answer No.''}, Mistral Response: \textit{`No. The average flying speed of a dove is around 24-35 miles per hour (40-56 kilometers per hour) for larger species like the Mourning Dove, but smaller species like the Laughing Dove can fly at speeds closer to 20 miles per hour (32 kilometers per hour). However, there is no species of dove that consistently reaches speeds above 35 miles per hour (56 kilometers per hour).''}
\bigskip

Again, we hypothesize that the root cause of this inconsistent behavior may be the non-preferred non-company reading of the \textit{People} and \textit{Location} categories, as well as in Study 3, which makes all interactions with the models more complicated and erroneous.

Finally, there are cases where the models fail to demonstrate knowledge of the non-company reading for \textit{People} and \textit{Location} entities (despite this knowledge being shown in Study 1 and evident in responses to questions in Study 3 by providing relevant and reasonable replies that contained the required information):

\bigskip

\noindent{}Prompt: \textit{``Is 1830 the date of birth of a person Philips? Answer with Yes or No and provide explanation.''}, Llama-3: \textit{``NO. Philips is a company, not a person. Philips is a Dutch multinational technology company founded in 1891 by Gerard Philips and his father Frederik Philips. It is not a person's date of birth.''}
\bigskip

\noindent{}Prompt: \textit{``Does an animal Jaguar have 35 to 50 miles per hour (56 to 80 kilometers per hour) speed? Answer ONLY with Yes or No. Answer based on your internal knowledge. Do not include any additional clarification. If you cannot answer this question, answer No.''}, Mistral: \textit{``Yes. A Jaguar can reach speeds between 50 to 80 kilometers per hour, which is approximately 31 to 49 miles per hour.''}

\section{Technical Details}
\label{app:technical_details}

None of our experiments require extensive computation or a large number of responses from LLMs.
Getting all responses from both GPT-3.5 and GPT-4o for all prompts coasted less than 30\$. 
For the Gemma and Mistral models, we used single H100 GPU,
and for Llama-3 and Mixtral two H100 GPUs. 
The responses for all studies were collected in less than 1 hour for each model.
We query GPT-4o and GPT-3.5 through the OpenAI API \footnote{\url{https://platform.openai.com/docs/api- 295 reference/introduction}}.
The open-sources models are available in the Huggingface Transformers library:
\begin{itemize}
    \item Gemma-7B: \url{https://huggingface.co/google/gemma-1.1-7b-it}
    \item Mistral-7B: \url{https://huggingface.co/mistralai/Mistral-7B-Instruct-v0.2}
    \item Mixtral-8x7B: \url{https://huggingface.co/mistralai/Mixtral-8x7B-Instruct-v0.1}
    \item Llama-3-70B: \url{https://huggingface.co/meta-llama/Meta-Llama-3-70B-Instruct}
\end{itemize}
Interactions with the models were conducted using the OpenAI Chat Completions API \footnote{\url{https://platform.openai.com/docs/libraries/python-library}}.  

To increase the reproducibility of our results, we fixed the seed value to 42. 
However, according to OpenAI's official documentation, the seed parameter is in beta, and determinism is not guaranteed (see \url{https://platform.openai.com/docs/api-reference/chat/create#chat-create-seed}). 
Therefore, we cannot fully guarantee the reproducibility of our results.
All models were checked for consistency by running the same prompt multiple times, with minimal variation observed across runs in all studies except for Study 4. Consequently, in Studies 1–3, we evaluate the results of a single run, while in Study 4, each prompt was run 5 times and the results were aggregated.
The temperature value was set to 1.0 in all experiments, and no system prompts were specified.

\begin{table*}[ht!]
    \centering
    \small
    \begin{tabularx}{\textwidth}{|X|}
    \hline
    \textbf{Study 1}\\ \hline
    Tell me about a company called <entity>.\\ \hline
    Tell me about an animal <animal-entity>.\\ \hline
    Tell me about a fruit <fruit-entity>.\\ \hline
    Tell me about a geographic location of <location-entity>.\\ \hline
    Tell me about a mythological character <myth-entity>.\\ \hline
    Tell me about a person <person-entity>.\\ \hline
    Tell me about a concept <abstract-entity>.\\ \hline
    Can <animal-entity> mean anything else but an animal? Answer only with Yes or No.\\ \hline
    Can <fruit-entity> mean anything else but a fruit? Answer only with Yes or No.\\ \hline
    Can <location-entity> mean anything else but a geographic location? Answer only with Yes or No.\\ \hline
    Can <myth-entity> mean anything else but a mythological character? Answer only with Yes or No.\\ \hline
    Can <person-entity> mean anything else but person? Answer only with Yes or No.\\ \hline
    Can <abstract-entity> mean anything else but an abstract concept? Answer only with Yes or No.\\ 
    \hline
    \textbf{Study 2} \\ \hline
    Group the following according to what they all have in common: <animal-entities>. Please provide an explanation. \\ \hline
    Group the following according to what they all have in common: <fruit-entities>. Please provide an explanation. \\ \hline
    Group the following according to what they all have in common: <location-entities>. Please provide an explanation. \\ \hline
    Group the following according to what they all have in common: <myth-entities>. Please provide an explanation. \\ \hline
    Group the following according to what they all have in common: <person-entities>. Please provide an explanation. \\ \hline
    Group the following according to what they all have in common: <abstract-entities>. Please provide an explanation. \\ \hline
    Group the following according to what they all have in common: <animal-entities>.\\ \hline
    Group the following according to what they all have in common: <fruit-entities>.\\ \hline
    Group the following according to what they all have in common: <location-entities>.\\ \hline
    Group the following according to what they all have in common: <myth-entities>.\\ \hline
    Group the following according to what they all have in common: <person-entities>.\\ \hline
    Group the following according to what they all have in common: <abstract-entities>.\\ \hline
    Group the following entities: <animal-entities>. Please provide an explanation. \\ \hline
    Group the following entities: <fruit-entities>. Please provide an explanation. \\ \hline
    Group the following entities: <location-entities>. Please provide an explanation. \\ \hline
    Group the following entities: <myth-entities>. Please provide an explanation. \\ \hline
    Group the following entities: <person-entities>. Please provide an explanation. \\ \hline
    Group the following entities: <abstract-entities>. Please provide an explanation. \\ \hline
    Group the following entities: <animal-entities>. \\ \hline
    Group the following entities: <fruit-entities>. \\ \hline
    Group the following entities: <location-entities>. \\ \hline
    Group the following entities: <myth-entities>. \\ \hline
    Group the following entities: <person-entities>. \\ \hline
    Group the following entities: <abstract-entities>. \\ \hline
    \textbf{Study 3} \\
    \hline
    Provide the founding year for <entity>. \\ \hline
    Provide the founding year for the company <entity>. \\ \hline
    Provide the speed for <animal-entity>. \\ \hline
    Provide the speed for an animal <animal-entity>. \\ \hline
    Provide the color for <fruit-entity>. \\ \hline
    Provide the color for a fruit <fruit-entity>. \\ \hline
    Provide the area in square meters for <location-entity>. \\ \hline
    Provide the area in square meters for a geographic location of <location-entity>. \\ \hline
    Provide the gender for <myth-entity>. \\ \hline
    Provide the gender for a mythological character <myth-entities>. \\ \hline
    Provide the date of birth for <person-entity>. \\ \hline
    Provide the date of birth for a person <person-entity>. \\ \hline
    Provide the level of abstractness for <abstract-entity>. \\ \hline
    Provide the level of abstractness for a concept <abstract-entity>. \\
    \hline
    \textbf{Study 4} \\ \hline
    Was there any company with the name (or part of the name) <entity> founded/established/launched/started in the year <retrieved-info>? Answer ONLY with Yes or No. If you cannot answer this question, answer No. \\\hline
    Does a concept <entity> has a <retrieved-info> level of abstractness? Answer ONLY with Yes or No. If you cannot answer this question, answer No. \\\hline
    Does a mythological character <entity> have a <retrieved-info> gender? Answer ONLY with Yes or No. If you cannot answer this question, answer No. \\ \hline
    Is there a geographic location <entity> with an approximate area of <retrieved-info>? Answer ONLY with Yes or No. If you cannot answer this question, answer No. \\ \hline
    Does a fruit <entity> have <retrieved-info> color? Answer ONLY with Yes or No. If you cannot answer this question, answer No. \\ \hline
    Does an animal <entity> have <retrieved-info> speed? Answer ONLY with Yes or No. If you cannot answer this question, answer No. \\ \hline
    Is <retrieved-info> the date of birth of a person <entity>? Answer ONLY with Yes or No. If you cannot answer this question, answer No. \\ \hline
    \end{tabularx}
    \caption{Prompts used in Studies 1-4.}
    \label{tab:all_prompts_study}
\end{table*}

\end{document}